%% file: ms.tex
\documentclass{article}
\usepackage{amsmath,graphicx,mlspconf}
\usepackage{xcolor}
\usepackage{amsfonts}
\usepackage{graphicx}
\usepackage{subcaption}
\usepackage{booktabs}
\usepackage{xurl}
\usepackage{orcidlink} 
\usepackage{hyperref}

\input{macros}

\copyrightnotice{979-8-3503-7225-0/24/\$31.00 {\copyright}2024 IEEE}

\toappear{2024 IEEE International Workshop on Machine Learning for Signal Processing, Sept.\ 22--25, 2024, London, UK}
\title{JOINT OPTIMIZATION OF PIECEWISE LINEAR ENSEMBLES}

\name{Matt Raymond$^1$\orcidlink{0000-0001-6824-8692} \qquad Angela Violi$^{1,2,3}$\orcidlink{0000-0001-9517-668X} \qquad Clayton Scott$^{1,4}$\orcidlink{0000-0002-0373-817X}}
\address{
    $^1$ Electrical Engineering and Computer Science,\; 
    $^2$ Mechanical Engineering, \\
    $^3$ Chemical Engineering,\; 
    $^4$ Statistics\\
    University of Michigan\\
    500 South State Street, Ann Arbor, 48109-2125, Michigan, USA \\
    \set{\texttt{mattrmd}, \texttt{avioli}, \texttt{clayscot}}@umich.edu
}

\begin{document}

\maketitle

\noindent\fbox{%
    \parbox{\linewidth}{%
        While preparing the PyPI package for general release, we found minor bugs in the penalty gradient computation and the validation set preprocessing that affected \joplen with a Laplacian + Frobenius-norm penalty and \joplen/GR with a CatBoost partitioner, respectively.
        Fixing these bugs provides the updated results shown in Figure~\ref{fig:regr} and Section~\ref{subsec:regclass}.
        The conclusions of the paper remain the same.
    }%
}

\begin{abstract}

Tree ensembles achieve state-of-the-art performance on numerous prediction tasks.
We propose {\textbf{J}oint} {\textbf{O}ptimization} of {\textbf{P}iecewise} {\textbf{L}inear} {\textbf{En}sembles} (\joplen), which jointly fits piecewise linear models at all leaf nodes of an existing tree ensemble.
In addition to enhancing the ensemble expressiveness, \joplen allows several common penalties, including sparsity-promoting and subspace-norms, to be applied to nonlinear prediction.
For example, \joplen with a nuclear norm penalty learns subspace-aligned functions.
Additionally, \joplen (combined with a Dirty LASSO penalty) is an effective feature selection method for nonlinear prediction in multitask learning.
Finally, we demonstrate the performance of \joplen on 153 regression and classification datasets and with a variety of penalties.
\joplen leads to improved prediction performance relative to not only standard random forest and boosted tree ensembles, but also other methods for enhancing tree ensembles.

\end{abstract}


\section{Introduction}
\label{sec:introduction}

Ensemble methods combine multiple prediction functions into a single prediction function.
A canonical example is a tree ensemble, where each tree ``partitions'' the feature space, and each leaf is a ``cell'' of the partition containing a simple (\eg constant) model.
Although neural networks (NN) have recently triumphed on structured data, they have struggled to outperform tree ensembles, such as gradient boosting (GB), across diverse tabular (\ie table-formatted) datasets~\cite{shwartz2021}.

Despite the longstanding success of tree ensembles, standard implementations are still limited, either by the piecewise constant fits at each cell, or by the greedy, suboptimal way in which the ensembles are trained.
To address these limitations, several improvement strategies have been proposed.
FASTEL uses backpropagation to optimize the parameters of an ensemble of smooth, piecewise constant trees~\cite{ibrahim2022}.
Global refinement (GR) jointly refits all constant leaves of a tree ensemble after first running a standard training algorithm, such a random forests or gradient boosting~\cite{ren2015}.
Partition-wise Linear Models also perform joint optimization, but learn linear functions on axis-aligned and equally-spaced step functions~\cite{oiwa2014}.
Linear Forests (LF) increase model expressiveness by replacing constant leaves with linear models~\cite{rodriguez2010}.

Because tree ensembles are nonlinear and greedily constructed, it has also been challenging to incorporate structure-promoting penalties through joint optimization.
RF feature selection typically relies on a heuristic related to the total impurity decrease associated to each feature.
This approach may underselect correlated features, and requires further heuristics to extend to the multitask setting.
GB feature selection uses a greedy approximation of the $\ell_1$ norm, which penalizes the addition of new features to each subsequent tree~\cite{xu2014}.
\bouts is a multitask extension that selects ``universal'' features (important for all tasks) and ``task-specific'' features by selecting features that maximize the minimum impurity decrease across all tasks~\cite{raymond2024}.
As such examples demonstrate, the incorporation of feature sparsity and similar structural objectives into tree ensembles has been limited by the greedy nature of ensemble construction.

We propose {\textbf{J}oint} {\textbf{O}ptimization} of {\textbf{P}iecewise} {\textbf{L}inear} {\textbf{En}sembles} (\joplen), an extension of global refinement (GR) that is applied to ensembles of piecewise linear functions.
\joplen both ameliorates greedy optimization and increases model flexibility by jointly optimizing a hyperplane in each cell (\eg leaf) in each partition (\eg tree) of an ensemble.
Besides improving performance, and unlike prior tree ensembles, \joplen is compatible with many standard structure-promoting regularizers, which provides a simple way to incorporate sparsity, subspace structure, and smoothness into nonlinear prediction.
We demonstrate this capability on 153 real-world and synthetic datasets; \joplen frequently outperforms existing methods for regression, binary classification, and multitask feature selection, including GB, RF, LF, CatBoost, soft decision trees (\fastel), and NNs.
Finally, we provide a GPU-accelerated implementation that is extensible to new loss functions, partitions, and regularizers.

\section{Methodology}
\label{sec:methodology}
We begin by describing \joplen in the single task setting, and subsequently extend it to multitask learning.

\subsection{Single-task \joplen}\label{subsec:st_joplen}
Let $\XX \doteq \R^d$ be the feature space and $\YY$ the output space.
For regression, $\YY \doteq \R$, and for binary classification, $\YY \doteq \{-1,1\}$.
Let $\set{(\x_\sampleitr,y_\sampleitr)}_{\sampleitr=1}^\nsamples$ be a training dataset where $\nsamples \in \N$, $\x_\sampleitr \in \XX$, and $y_\sampleitr \in \YY$.
A model class $\FF$ is a set of functions $f: \XX \to \R$.
Given $\FF$, the goal of single-task supervised learning is to find an $f \in \FF$ that accurately maps feature vectors to outputs.
For regression and binary classification, predictions are made using $f(\x)$ and $\sign(f(\x))$, respectively.

\joplen is an instance of (regularized) empirical risk minimization (ERM).
In ERM, a prediction function $\hf$ is a solution of 
\begin{equation}
  \argmin_{f \in \FF} \sum_{\sampleitr=1}^\nsamples \ell(y_\sampleitr, f(\x_\sampleitr)) + R(f) \;,
\end{equation}
where $R\colon \FF \to [0,\infty)$ is a regularization term (penalty), and $\ell\colon \YY \times \R \to [0,\infty)$ is a loss function.
We take $\ell$ to be the squared error loss for regression, and the logistic loss for binary classification.

We focus on the setting where $f \in \FF$ is an additive ensemble $f \doteq \sum_{\partitr = 1}^\nparts f_\partitr$, where $f_p$ is defined piecewise on a partition (\eg a decision tree), and there are $\nparts \in \N$ partitions.
The partitions are fixed in advance, and may be obtained by running a standard implementation of a tree ensemble model.
Further, we assume that $f_\partitr(\x) \doteq \sum_{\cellitr = 1}^\ncells (\ip{\w_{\partitr,\cellitr}}{\x} + b_{\partitr,\cellitr}) \phi(\x; \partitr,\cellitr)$.
Here $\ncells\in \N$ is the number of ``cells'' in the partition (\eg tree), $\w_{\partitr,\cellitr} \in \R^{\nfeatures}$ is a weight vector, $b_{\partitr,\cellitr} \in \R$ is a bias term, and $\phi(\cdot ; \partitr,\cellitr)\colon \XX \to \set{0,1}$ indicates whether data point $\x$ is in cell $\cellitr$ of partition $\partitr$~\cite{ren2015} (\eg indicating the decision regions of a tree).
Then, we denote a piecewise linear ensemble as
\begin{equation}
  f(\x; \W, \b, \phi) = \sum_{\partitr = 1}^\nparts \sum_{\cellitr = 1}^\ncells (\ip{\w_{\partitr,\cellitr}}{\x} + b_{\partitr,\cellitr}) \phi(\x; \partitr,\cellitr) \;,
\end{equation}
where $C \doteq \sum_{p = 1}^\nparts \ncells$ is the total number of cells, $\W \in \R^{\nfeatures \times C}$ is the matrix of all weight vectors, $\b \in \R^{C}$ is a vector of all bias terms, and $\square_{p,c}$ is the component of $\square$ associated with the $c$th cell of partition $p$.
\joplen is then defined by the solution of
\begin{equation}
  \argmin_{\W, \b} \frac{1}{\nsamples} \sum_{\sampleitr = 1}^\nsamples \ell\left(y_\sampleitr, f(\x_\sampleitr; \W, \b, \phi)\right) + R(\W, \b; \x_\sampleitr) \;,
\end{equation}
where $R\colon \R^{d\times C}\times\R^C\times\XX \to [0,\infty)$ is a regularization term (penalty).
The notation $R(\W, \b; \x_\sampleitr)$ indicates that $R$ may depend on $\x_\sampleitr$, but that only $\W$ and $\b$ will be penalized.

\joplen has attractive optimization properties by construction.
For example, a convex loss and regularization term will render the overall objective convex.
We optimize \joplen using Nesterov's accelerated gradient method, and we use a proximal operator for any non-smooth penalties.

\joplen easily incorporates existing penalties, and provides a straightforward framework for extending well-known penalties for linear models to a nonlinear setting.
Notable examples include sparsity-promoting matrix norms (\eg $\ell_{p,q}$-norms~\cite{obozinski2006, hu2017} and GrOWL~\cite{oswal2016}) and subspace norms (nuclear, Ky Fan%
, and Schatten $p$-norms)~\cite{fessler2024}.
We demonstrate two such standard penalties:
$\ell_{p, 1}$-norms and the nuclear norm.

\subsubsection{Single-task $\ell_{p,1}$-norm}\label{subsubsec:st_l21}

Because we use linear leaf nodes, we can use sparsity-promoting penalties to preform feature selection. 
To learn a consistent sparsity pattern across all linear terms, we realize $R$ as an $\ell_{p,1}$ sparsity-promoting group norm (for $p\geq 1$), which leads to a row-sparse solution for $\W$.
Specifically,
\begin{equation}
\begin{split}
    R_{p,1}(\W, \b; \x_\sampleitr) &\doteq \lambda \|\W\|_{p,1}\\
    &\doteq \lambda \sum_{\featureitr = 1}^{\nfeatures} \left(\sum_{\partitr, \cellitr = 1}^{\nparts, \ncells} |\w_{\partitr, \cellitr, \featureitr}|^p\right)^{1/p} \; ,
\end{split}
\end{equation}
where $\w_{\partitr, \cellitr, \featureitr}$ is the $i$th component of the vector $\w_{\partitr, \cellitr}$.
Concretely, we use the $\ell_{2,1}$ norm, which has a straightforward proximal operator~\cite[Eq.~6.8, p.~136]{parikh2014}.

\subsubsection{Single-task nuclear norm}\label{subsubsec:st_nn}

Given a singular value decomposition $\W \doteq \mathbf{U\Sigma V}^\mathsf{H}$ for the conjugate transpose $\mathsf{H}$, the nuclear norm is defined as 
\begin{equation}
    R_{\text{nn}}(\W, \b; \x_\sampleitr) \doteq \lambda \|\W\|_* \doteq \lambda \hspace{-0.3cm} \sum_{i=1}^{\min(C, \nfeatures)} |\mathbf{\Sigma}_{i,i}| \;.
\end{equation}
When applied to \joplen's weights, the nuclear norm penalty empirically functions similarly to a group-norm penalty.
While group-norms promote sparse solutions in an axis-aligned subspace, the nuclear norm apparently promotes ``sparse'' solutions along the ``axes'' of a data-defined subspace.
This norm also has a straightforward proximal operator using singular value thresholding~\cite[Eq.~6.13]{parikh2014}.

\subsection{Multitask \joplen}\label{subsec:mt_joplen}

Suppose we have $\ntasks \in \N$ datasets indexed by $\taskitr \in \set{1,...,\ntasks}$: $\set{(\x^\taskitr_\sampleitr, y^\taskitr_\sampleitr)}^{\nsamples^\taskitr}_{\sampleitr = 1}$.
Multitask \joplen is nearly identical to the single-task case, but includes an additional sum over tasks.
Let $\W^\taskitr \in \R^{\nfeatures \times C^t}$ for a task-specific number of cells $C^t\in \N$, $\b^\taskitr \in \R^{C^t}, \phi^\taskitr\colon \XX \to \set{0,1}$, and $\gamma^\taskitr \in (0,\infty)$ be the $t$-th task's weight matrix, bias vector, partitioning function, and relative weight.
For notational simplicity, define $\W \doteq [\W^1 \cdots \W^\ntasks]$, $\B \doteq [\b^1 \cdots \b^\ntasks]$, and $\vgamma \doteq [\gamma^1 \cdots \gamma^\ntasks]^\top$ for the transpose $\top$.
Then multitask \joplen is defined as
\begin{equation}
    \begin{split}
        \argmin_{\W,\B} \sum_{\taskitr=1}^\ntasks \frac{\taskweight^\taskitr}{\nsamples^\taskitr} \sum_{\sampleitr = 1}^{\nsamples^\taskitr} &\ell\left(y_\sampleitr^\taskitr, f(\x_\sampleitr^\taskitr; \W^\taskitr, \b^\taskitr, \phi^\taskitr)\right) \\ &+ R(\W, \B; \vgamma, \x_\sampleitr) \; ,
    \end{split}
\end{equation}
and can be optimized using the same approach as in the single-task setting.

\subsubsection{Dirty LASSO}\label{subsubsec:mt_dl}

\joplen with an extended Dirty LASSO (DL)~\cite{jalali2010} penalty performs feature selection for nonlinear multitask learning.
Suppose that we know \emph{a priori} that some features are important for all $T$ tasks, some features are important for only some tasks, and all other features are irrelevant.
Then we can apply a \joplen extension of DL to perform ``common'' and ``task-specific'' feature selection.

DL decomposes the weight matrix $\W^\taskitr \doteq \C^\taskitr + \T^\taskitr$, encouraging $\C \doteq [\C^1 \cdots \C^\ntasks]$ to be row-sparse (common features) and each $\T^\taskitr$ to be individually row-sparse (task-specific features), with potentially different sparsity patterns for each  $\T^\taskitr$.
Given penalty weights $\lambda_\C, \lambda_\T \in [0,\infty)$, \joplen DL is the solution of
\begin{equation}
\begin{split}
  \argmin_{\C, \T, \B} \sum_{\taskitr=1}^\ntasks \Bigg[ \frac{\taskweight^\taskitr }{\nsamples^\taskitr}\sum_{\sampleitr = 1}^{\nsamples^\taskitr} &\ell\left(y_\sampleitr^\taskitr, f(\x_\sampleitr^\taskitr; \C^\taskitr + \T^\taskitr, \b^t, \phi^\taskitr)\right) \\&+ \gamma^\taskitr\lambda_\T\|\T^\taskitr\|_{2,1} \Bigg] + \lambda_\C\|\C\|_{2,1} \; ,
\end{split}
\end{equation}
a combination of LASSO and Group LASSO penalties~\cite{jalali2010}.

\subsection{Laplacian regularization}\label{subsec:laplacian}

The na\"ive use of piecewise-linear functions may lead to pathological discontinuities at cell boundaries.
Thus, we utilize graph Laplacian regularization~\cite{zhu2003} to force nearby points to have similar values.
Using the standard graph Laplacian,
\begin{equation}
    R_{\Ell}(\W, \b; \x_\sampleitr) \doteq \frac{\lambda}{2}\sum_{i, j = 1}^{\nsamples} \K(\x_{i}, \x_{j})(\hat{y}_{i} - \hat{y}_{j})^2 \; ,
\end{equation}
where $\hat{y}_i \doteq f(\x_i ; \W, \b, \phi)$ is the model prediction for a feature vector $\x_i$ and $\K\colon \XX \times \XX \to \R$ is a distance-based kernel.
In this paper, $\K$ is a Gaussian radial basis function.
Laplacian regularization can be na\"ively applied to each task in the multitask setting as well.

\section{Experiments}
\label{sec:experiments}

We evaluate \joplen in multiple regression and classification settings with several regularization terms.

\subsection{Single-task regression and binary classification}\label{subsec:regclass}
We evaluate \joplen's predictive performance on the ``Penn Machine Learning Benchmark'' (PMLB) of 284 regression, binary classification, and multiclass classification tasks~\cite{romano2021}.
For simplicity, we focus on regression and binary classification.
Some methods (\eg GB, \joplen) only handle real-valued features, so we drop all other features for such methods.
We ignore datasets that have no real-valued features, or are too large for our GPU.
This leaves 90 regression and 60 binary classification datasets.
For each dataset, we perform a random 0.8/0.1/0.1 train/validation/test split.

To facilitate comparisons with previous methods~\cite{ren2015}, we jointly optimize partitions created using Gradient Boosted trees~\cite{friedman1999}, Random Forests~\cite{breiman1996a}, and CatBoost~\cite{prokhorenkova2018}.
We also evaluate \joplen with random Voronoi ensembles (Vor).
Each partition $p$ in a Voronoi ensemble is created by sampling $\ncells$ data points uniformly at random and creating Voronoi cells from these data points.
This is to provide context for the performance of \joplen in Section~\ref{subsec:mtselection} and Figure~\ref{fig:mtperf}~\bb.

To demonstrate \joplen's efficacy, we benchmark several alternative models: Gradient Boosted trees (GB), Random Forests (RF)~\cite{pedragosa2011}, Linear Forests (LF)~\cite{cerliani2022, rodriguez2010}, and CatBoost (CB)~\cite{prokhorenkova2018}; Linear/Logistic Ridge Regression~\cite{pedragosa2011}; a feedforward neural network (NN)~\cite{tensorflow} and a NN for tabular data (TabNet)~\cite{arik2020,pytorchtabnet}; a differentiable tree ensemble (\fastel)~\cite{ibrahim2022}; and global refinement (GR)~\cite{ren2015}.
GB, RF, LF, and CB are baseline ensembles for demonstrating the performance improvement from joint optimization.
CB utilizes categorical features and corrects a bias in GB's gradient step~\cite{prokhorenkova2018}.
\fastel and TabNet provide a comparison between joint optimization, fully-differentiable ensembles, and tabular deep learning.
We evaluate CatBoost both with and without (NC) categorical features.
We exclude \fastel from the classification experiments because the code does not include classification losses.
Global refinement (GR) jointly optimizes piecewise-constant tree ensembles.
We exclude the global pruning step from \cite{ren2015} to more clearly demonstrate the direct performance improvement of \joplen's piecewise-linear approach.
These approaches provide a thorough comparison between \joplen and existing methods.

As in \cite{ren2015}, we use the squared Frobenius norm of leaf nodes for both piecewise-constant and piecewise-linear ensembles.
We also evaluate a combined Laplacian-Frobenius penalty for piecewise-linear ensembles (Section \ref{subsec:laplacian}).

We optimize model hyperparameters using Ax~\cite{bakshy2018}, which combines Bayesian and bandit optimization for continuous and discrete parameters.
Ax uses 50 training/validation trials to model the validation set's empirical risk as a function of hyperparameters, and we use the optimal hyperparameters to evaluate model performance on the test set.
Hyperparameter ranges are documented in our code.

To facilitate plotting, we normalize the mean squared error (MSE) of each regression dataset by the MSE achieved by predicting the training mean and call this the ``normalized'' MSE.
The 0/1 loss does not require normalization.

\begin{figure}[!htb]
    \centering
    \includegraphics[width=\linewidth]{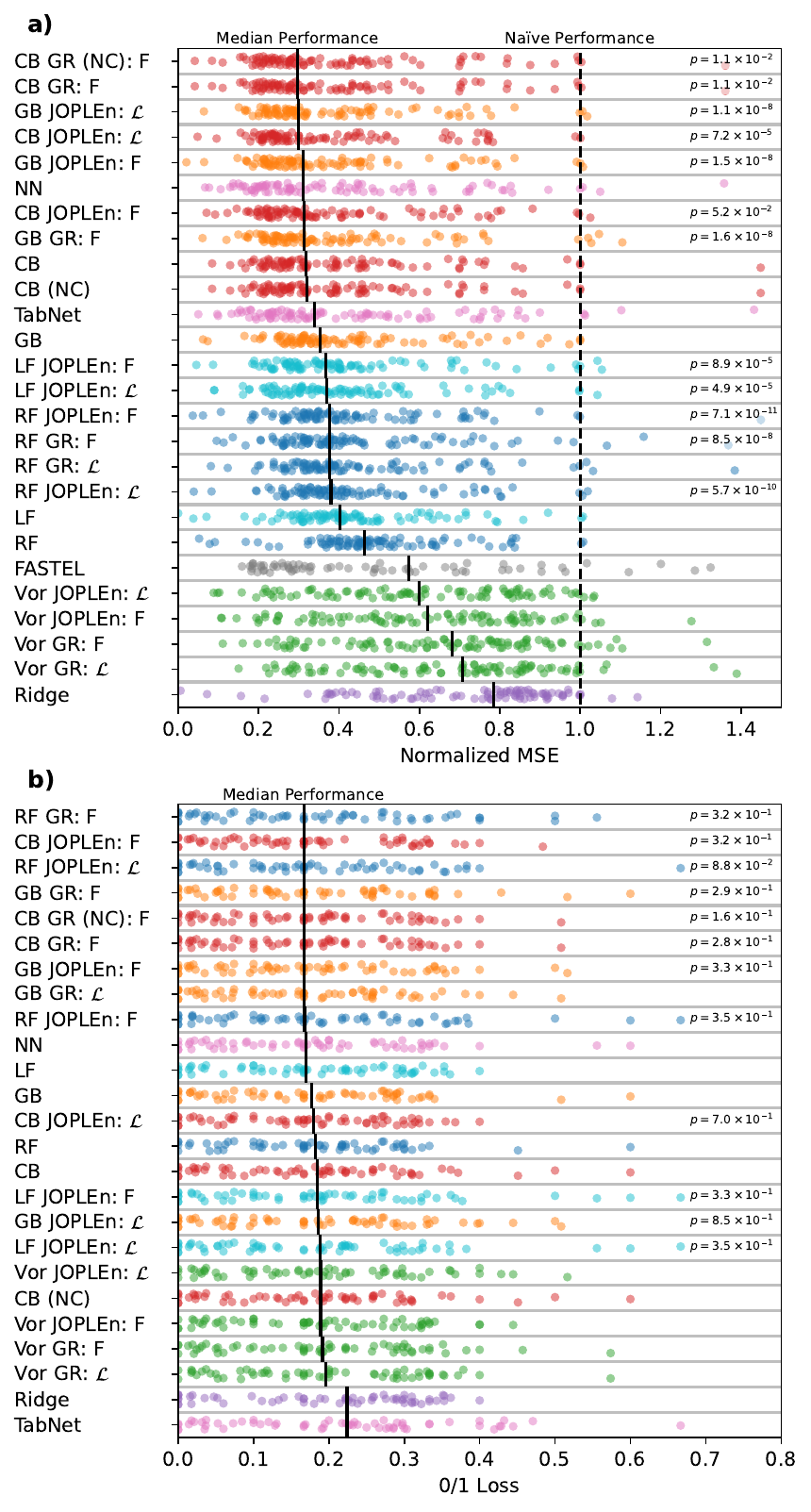}
    \caption{
        Each point is one PMLB data set.
        Similar models are grouped by color.
        ``$\mathcal{L}$'' and ``F'' indicate Laplacian + Frobenius (F) norm and F norm regularization.
        ``NC'' indicates CatBoost without categorical features.
        \aa shows the normalized MSE on regression datasets (truncated at 1.5).
        The dotted line indicates na\"ive performance.
        Right-hand $p$-values compare the refitting method and with the original ensemble.
        \bb shows the 0/1 loss for classification datasets.
        For both plots, the black line indicates the median performance over all datasets.
    }
    \label{fig:regr}
\end{figure}

Similar to \cite{ren2015}, Figure~\ref{fig:regr}~\aa shows that \joplen and GR significantly improve the regression performance of GB and RF, as determined by a one-sided Wilcoxon signed-rank test.
\joplen also significantly improves LF and CB.
Indeed, \joplen CB $\mathcal{L}$ outperforms all regression approaches including CB ($p=7.2\times 10^{-5}$), despite dropping all categorical features.
Based on median performance, CB GR (NC), CB GR, and GB \joplen $\mathcal{L}$ appear to outperform \joplen CB $\mathcal{L}$.
However, \joplen CB $\mathcal{L}$ outperforms them with $p=5.6\times10^{-3}$, $p=4.9\times10^{-3}$, and $p=8.0\times10^{-2}$ due to their heavier tail performance.

Although \joplen also improves performance for binary classification, the difference is less pronounced ($p=8.8\times 10^{-2}$ for \joplen RF $\mathcal{L}$).
This decreased significance may be caused by the sample size (60 \vs 90 datasets), or the fact that the 0/1 loss only considers the sign of the prediction.

Interestingly, we observe that the partitioning method may significantly affect performance.
In the regression setting, CB, GB, LF, RF, and Voronoi model performances rank in this order.
However, there is no significant difference in the classification setting.
We suspect that there may be other partitioning methods that lead to further performance increases.

Finally, we note that CB \joplen $\mathcal{L}$ outperforms the NNs and \fastel in general, achieving $p$-values of $3.3\times10^{-4}$ and $1.5\times10^{-1}$ for TabNet and the feedforward network.
The NNs and \fastel perform well on many datasets, but have a high median loss because some datasets perform quite poorly.
This is particularly true for \fastel.

\subsection{Single-task nuclear norm}\label{subsec:nuclear}

\begin{figure}
    \centering
    \includegraphics[width=\linewidth]{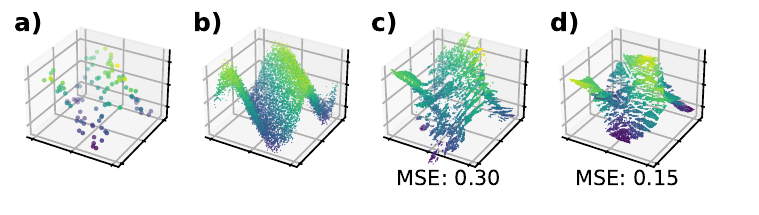}
    \caption{
        \aa and \bb show the training and testing sets of a function that lies along a feature subspace.
        \cc shows linear \joplen's prediction using a Frobenius norm penalty, and \dd shows the prediction using the nuclear norm penalty.
        The mean squared error (MSE) is reported below each method.
    }
    \label{fig:nn}
\end{figure}

In Figure~\ref{fig:nn} we demonstrate the effect of the nuclear norm on synthetic data.
We draw 10,000 samples (100 are training data) uniformly on the square interval $[-1,1]^2$, and define $y_\sampleitr \doteq f(\x_{\sampleitr})$ where $f(\x) \doteq \sin(\pi (x_1 + x_2)) + \varepsilon$ for an input feature $\x$ and $\varepsilon \sim \mathcal{N}(0,0.2)$.
Here, we use random Voronoi ensembles and fix the number of partitions and cells to simplify visualization, and manually tune the penalty weights.

Notably, the nuclear norm causes the model to align its piecewise linear functions along the diagonal, forming a ``consensus'' across all linear models. 
By contrast, the Frobenius norm penalty allows each leaf to be optimized independently, and thus suffers from degraded performance.

\subsection{Multitask Dirty LASSO}\label{subsec:mtselection}
Next, we show that \joplen Dirty LASSO (DL) can achieve superior sparsity to the DL~\cite{jalali2010}.

We selected two multitask regression datasets from the literature: NanoChem~\cite{raymond2024} and SARCOS~\cite{vijayakumar2005}.
NanoChem is a group of 7 small molecule, nanoparticle, and protein datasets for evaluating multitask feature selection performance~\cite{raymond2024}.
These datasets have 1,205 features and 127 to 11,079 samples.
The response variables are small molecule boiling point (1), Henry's constant (2), logP (3) and melting point (4); nanoparticle logP (5) and zeta potential (6); and protein solubility (7).
SARCOS is a 7-task dataset that models the dynamics of a robotic arm~\cite{vijayakumar2005}, with 27 features and 48,933 samples for each task.
These datasets were also split into train, validation, and test sets using a 0.8/0.1/0.1 ratio.

Using \joplen DL with tree ensembles will provide biased sparsity estimates; features are used to create each tree (not penalized) and then to train each linear leaf (penalized).
In this case, simply analyzing the leaf weights will underestimate the number of features used by the \joplen model.
We avoid this issue by using random Voronoi ensembles.

All penalties were manually tuned so that all methods achieve a similar test loss for each task.
Not all datasets are equally challenging, so we manually tune the $\gamma^\taskitr$ parameter for \joplen DL and \bouts.
No equivalent parameters exist in the community implementation of DL~\cite{janati2019}.

\begin{figure}[htb]
    \centering
    \includegraphics[width=\linewidth]{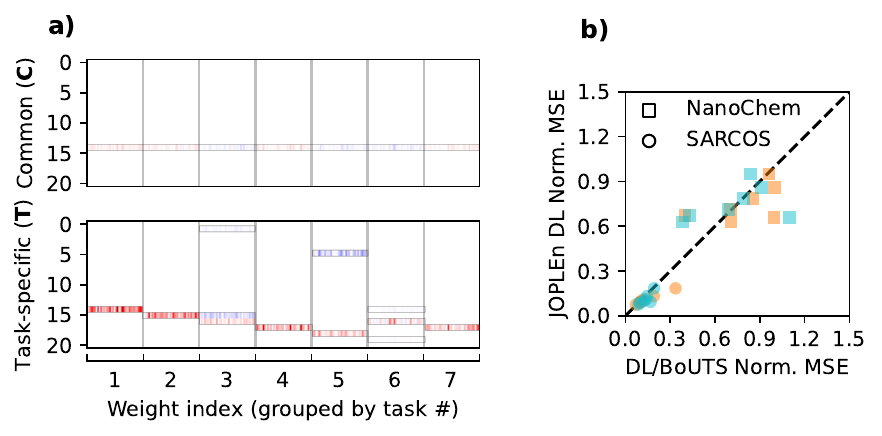}
    \caption{
        \aa demonstrates that \joplen DL learns common and task-specific features from the SARCOS dataset.
        The $x$-axis represents cells grouped by task, and the $y$-axis indicates the associated input feature.
        Blue indicates negative weights, red indicates positive weights, white is zeros.
        \bb shows regression performance, with equal performance on the diagonal.
        \joplen's performance is equal to or greater than that of DL (orange) on most tasks, and similar to that of \bouts (teal).
    }
    \label{fig:mtperf}
\end{figure}

\begin{figure}[htb]
    \centering
    \includegraphics[width=\linewidth]{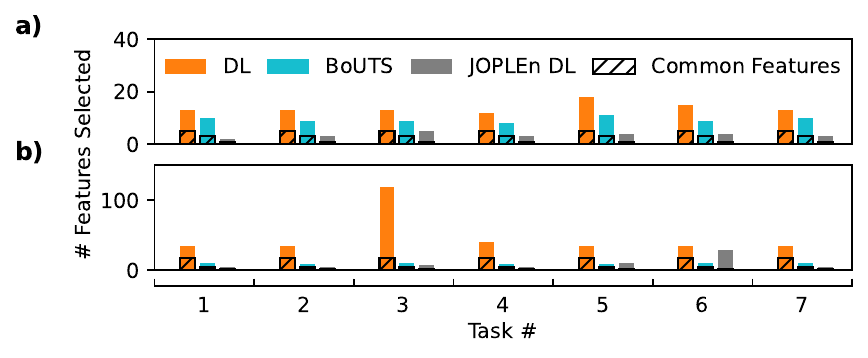}
    \caption{
        The number of common and task-specific features selected by DL, \bouts, and \joplen DL for the \aa SARCOS and \bb NanoChem datasets (fewer is better).
    }
    \label{fig:feat_sel}
\end{figure}

We find that \joplen DL and \bouts select significantly sparser feature sets than DL does, with \joplen providing the sparsest sets.
Figure~\ref{fig:mtperf}~\aa shows that joint optimization shares penalties across ensemble terms, providing structured sparsity.
Further, \joplen selects far fewer features than DL (Fig.~\ref{fig:feat_sel}) while achieving similar or superior performance on all but one task (small molecule logP (3), Fig.~\ref{fig:mtperf}~\bb).
This is likely because of the significant disparity in the number of features selected for this task (102 for DL, \vs 6 for \joplen).

\section{Conclusions}
\label{sec:conclusions}

The \joplen framework enables many existing penalties for linear methods (\eg single- and multitask feature selection) to be incorporated into nonlinear methods.

We find that \joplen beats global refinement (GR) and significantly outperforms other state-of-the-art methods on real-world and synthetic datasets.
We demonstrate that Laplacian; Frobenius, nuclear, $\ell_{p,1}$-norm; and Dirty LASSO regulation have straightforward extensions using \joplen and improve model performance.
Empirically, \joplen shines when the response variable is a nonlinear function of input features with structured sparsity, such as multitask feature selection.
Such settings cannot be modeled using linear approaches such as LASSO, and \joplen achieves similar or higher feature sparsity and performance than \bouts (Figure~\ref{fig:mtperf}~\bb) while using suboptimal partitions (Figure~\ref{fig:regr}).
Future work may use \bouts as a partitioner for \joplen to improve performance while maintaining a high degree of sparsity.
Combining \joplen with global pruning, developing new partitioning methods, and providing support for categorical features may also lead to further improvements.
Overall, these results suggest that \joplen is a promising approach to improving performance on tabular datasets.

We anticipate that \joplen will improve regression and classification performance across many fields.
Additionally, we expect that \joplen's piecewise linear formulation will lead to improved interpretability through increased feature sparsity.
Finally, because \joplen allows the straightforward transfer of linear penalties to the nonlinear setting, we anticipate that it will greatly simplify the implementation of nonlinear penalized optimization problems (\eg subspace-aligned nonlinear regression via the nuclear norm, Figure~\ref{fig:nn}).

\section{Code availability}

A JAX (CPU and GPU compatible) version of \joplen is available on PyPI as \texttt{joplen} with linked source code.
The data, \joplen implementation, and evaluation/plotting code for this paper is available at \url{https://gitlab.eecs.umich.edu/mattrmd-public/joplen-repositories/joplen-mlsp2024}.

\bibliographystyle{IEEEbib}
\bibliography{refs}

\end{document}

%% file: macros.tex
\usepackage{xspace}

\newcommand*{\joplen}{JOPLEn\xspace}

\DeclareMathAlphabet\mathsf{OT1}{lcmss}{m}{n}

\newcommand*{\fastel}{FASTEL\xspace}
\newcommand*{\bouts}{BoUTS\xspace}

\newcommand*{\ie}{\textit{i.e.,}\xspace}

\newcommand*{\eg}{\textit{e.g.,}\xspace}

\newcommand*{\vs}{\textit{vs.}\xspace}

\renewcommand{\vec}[1]{\boldsymbol{#1}}
\newcommand*{\set}[1]{\{#1\}}
\newcommand*{\N}{\mathbb{N}}
\newcommand*{\R}{\mathbb{R}}
\newcommand*{\ncells}{{C_{\partitr}}}
\newcommand*{\nparts}{P}
\newcommand*{\nfeatures}{d}
\newcommand*{\nsamples}{N}
\newcommand*{\ntasks}{T}

\newcommand*{\partitr}{p}
\newcommand*{\cellitr}{c}
\newcommand*{\sampleitr}{n}
\newcommand*{\featureitr}{i}
\newcommand*{\taskitr}{t}

\newcommand*{\W}{\mathbf{W}}

\newcommand*{\B}{\mathbf{B}}
\newcommand*{\C}{\mathbf{C}}
\newcommand*{\T}{\mathbf{T}}

\newcommand*{\w}{\vec{w}}
\renewcommand*{\b}{\vec{b}}

\newcommand*{\vgamma}{\vec{\gamma}}

\newcommand*{\K}{\mathbf{K}}

\newcommand*{\x}{\vec{x}}

\newcommand*{\Ell}{\mathcal{L}}

\newcommand*{\taskweight}{\gamma}
\newcommand{\XX}{\mathcal{X}}
\newcommand{\YY}{\mathcal{Y}}
\newcommand*{\FF}{\mathcal{F}}

\newcommand*{\ip}[2]{\langle #1, #2 \rangle}
\DeclareMathOperator*{\sign}{\text{sign}}

\newcommand*{\hf}{\hat{f}}

\DeclareMathOperator*{\argmin}{\arg\min}

\renewcommand*{\aa}{\textbf{a)}\xspace}
\newcommand*{\bb}{\textbf{b)}\xspace}
\newcommand*{\cc}{\textbf{c)}\xspace}
\newcommand*{\dd}{\textbf{d)}\xspace}